\documentclass[12pt,a4paper]{article} 
\usepackage{amsmath,amssymb}
\usepackage{mathptmx,url} 
\usepackage{amsfonts,xcolor,longtable,pbox}
\usepackage{graphicx}
\usepackage{subfigure}
\usepackage{algorithm2e}
\usepackage{abstract}

\newcommand{\Rr}{{\mathbb R}}
\title{Active Contour Models for Manifold Valued Image Segmentation}
\author{Sumukh Bansal\footnote{sumukh\_bansal@daiict.ac.in} \and Aditya Tatu\footnote{Queries and comments to be addressed to aditya.tatu@gmail.com}}
\begin{document}
\maketitle




\begin{abstract}
Image segmentation is the process of partitioning an image into different regions or groups based on some characteristics like color, texture, motion or shape etc. Active contours are a popular variational method for object segmentation in images, in which the user initializes a contour which evolves in order to optimize an objective function designed such that the desired object boundary is the optimal solution.\\
Recently, imaging modalities that produce Manifold valued images have come up, for example, DT-MRI images, vector fields. The traditional active contour model does not work on such images.\\
In this paper, we generalize the active contour model to work on Manifold valued images. As expected, our algorithm detects regions with similar Manifold values in the image. Our algorithm also produces expected results on usual gray-scale images, since these are nothing but trivial examples of Manifold valued images. As another application of our general active contour model, we perform texture segmentation on gray-scale images by first creating an appropriate Manifold valued image. We demonstrate segmentation results for manifold valued images and texture images.
\end{abstract}
\section{Introduction}
Image segmentation approaches are based on a characteristic property that defines the region of interest to be segmented out of the image, for example color, texture, motion, shape and/or others. \\
Image segmentation approaches can broadly be categorized into edge based or region based ones. There are various approaches in both categories based on intensity, color, texture and motion using statistical and geometrical framework \cite{cremers}.
Active contour model is a popular segmentation approach, which has both edge based \cite{kimmel} and region based \cite{chan} versions. \\
Active contours, also known as `snakes', are based on evolving an initial contour towards the boundary of an object to be detected \cite{kass}. Usually, this evolution equation is a gradient descent for minimizing an appropriate energy functional.\\
In the Geodesic active contours model \cite{kimmel}, the energy functional is written as a length functional with a modified metric such that the minimum, i.e. curve of minimum length corresponds to the object's boundary. A region based approach was proposed by Chan and Vese \cite{chan} where the energy functional was not based on edges. While the traditional approach works with parametric representation of curve, a simpler and efficient representation -the level set approach was introduced by Osher and Sethian \cite{osher88}.  \\
Grayscale images can be modeled as functions $I : \Omega \rightarrow \mathbb{R}$ where $\Omega \subseteq \mathbb{R}^{2}$ is the image domain. These days, there are imaging modalities and other data that can be modeled as a function $I : \Omega \rightarrow M$, where $M$ is a Riemannian manifold. We call such images Manifold valued images (henceforth written as MVI). Examples are DT-MRI images where $M=PD(3)$, the set of $3\times 3$ symmetric positive definite matrices, optical flow field images $M=\mathbb{R}^2$ or wind velocity data where  $M=\mathbb{R}^3$. The standard active contour methods will not work on MVIs. In this paper we generalize the active contour model to work on MVI. We adapt both, Geodesic active contours and Chan-Vese active contour models to work on MVIs. Since $\Rr$ is also an example of a manifold, our model also works on gray-scale images (modeled as $f:\Omega \rightarrow \Rr$).\\
As another application, we pose the texture segmentation problem as a MVI segmentation problem, as follows. Since covariance matrices are often used to characterize textures, we define a covariance matrix of gray-valued images in a neighborhood on every pixel of the given texture image. Covariance matrices are symmetric positive definite (actually semi-definite) matrices of size $n \times n$, the set of these matrices forms a manifold, denoted by $PD(n)$. Our algorithm is then able to identify regions of similar covariance matrices, thus segmenting textures in the image. In the rest of the paper $||\cdot||$ denotes the usual Euclidean norm, $|\cdot|$ denoted absolute value of a real number and symbols for other norms used have been defined where used for the first time.
\subsection{Related Work}
{There have been related works on DT-MRI image segmentation using active contours, for example work by Lisa Jonasson \cite{jonasson_geometricflow},\cite{jonasson_levelset}, where the surface $S$ is evolved according to
\begin{equation}
\frac{\partial S}{\partial t} = (F+H)\hat{n},
\end{equation}
where $F$ is a speed term proportional to the similarity of the Diffusion tensors of adjacent pixels along the normal direction to the surface, $H$ is a curvature based regularization term and $\hat{n}$ is the unit normal to the surface $S$ being evolved. The similarity measure between two tensors $\mathbf{T}_1$ and $\mathbf{T}_2$ used in the above mentioned paper is the Normalized Tensor Scalar Product (NTSP):
\[
NTSP(\mathbf{T}_1,\mathbf{T}_2) = \frac{Trace(\mathbf{T}_1\ \mathbf{T}_2)}{Trace(\mathbf{T}_1)\ Trace(\mathbf{T}_2)}
\]
A more Chan-Vese type active contour model was proposed by  Wang and Vemuri \cite{wang05} for DT-MRI segmentation, where they define the following energy function on the set of curves $C$:
\begin{equation}
E(C,\mathbf{T_1},\mathbf{T_2}) = \int_{R} ||\mathbf{T}(x) - \mathbf{T_1}||^2_F \ dx + \int_{R^c} ||\mathbf{T}(x) - \mathbf{T_2}||^2_F \ dx ,
\end{equation}
where $\mathbf{T}(x)$ is the tensor defined at $x$, $R$ ($R^c$) is the interior (exterior) of the curve $C$, and 
\begin{align}
\mathbf{T_1} = &\arg\min_{\mu}\left(\int_{R} ||\mu - \mathbf{T}(x)||^2_F\ dx\right),\\
\mathbf{T_2} = & \arg\min_{\mu}\left(\int_{R^c} ||\mu - \mathbf{T}(x)||^2_F\ dx\right)
\end{align}
are the mean tensors in the interior and exterior of the curve $C$. They use the Frobenius norm ($||.||_F$) in their computations. 
In \cite{Lenglet06}, the authors use Geodesic active contour to segment DT-MRI images. They use the Euclidean, KL-divergence based and the geodesic distance metric to induce an edge-stopping term. They conclude that the geodesic distance based metric yields the best segmentation results.
In \cite{Guo08}, the authors use a front propagation scheme to segment DT-MRI images.\\

Sagiv, Sochen and Zeevi\cite{sagiv06} propose a Chan-Vese type active contour model for texture segmentation. Instead of defining the model on intensity-valued texture image, they first compute responses to a manually selected set of Gabor filters $h_{mn}$:
\[
W_{mn}(x,y) = I(x,y) * h_{mn}(x,y),\ m = 1,\ldots,M,\ n = 1,\ldots, N.
\]
The scale, orientation and the response which are maximal are included into the feature at every point of the image. This feature space is a $2$-D manifold embedded in $\Rr^N$ (they use $N = 7$). Chan-Vese active contour is then run on an image where every pixel maps to a point in this feature space. These features can also be used to induce a metric on the $2$-D manifold. Lee et.al. \cite{Lee05} use this metric to define a stopping term for Geodesic active contours that segments out texture.\\
One of the other commonly used feature for characterizing textures is the covariance matrix. In \cite{Wildenauer07}, the authors use covariance matrices of the intensity and first order Gaussian derivatives at every pixel over a $5 \times 5$ neighborhood as a feature and segment textures using multi-scale Graph cuts. Donoser and Bischof \cite{Donoser08} use covariance matrices of intensity and first and second derivatives of intensity values as features to characterize textures. They use the manifold distance on the set of positive semi-definite matrices to extend the ROI-SEG \cite{Donoser07} clustering algorithm to segment textures.}
Our contribution is twofold: We give a general active contour model that can be used with intensity valued, vector valued, manifold valued images and as a computer vision application we show how this algorithm can be used to segment textures. \\
The paper is organized as follows.
In the next section we provide a brief background on active contours, manifold valued images and cite other work on manifold valued data. In section \ref{sec:MVIsegmentation}, we explain our proposed active contour model for MVI segmentation, followed by implementation details and results.
\begin{figure}
    \centering
    \subfigure[]
    {
        \includegraphics[width=1 in]{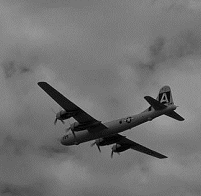}
        \label{fig:}
    }
    \subfigure[]
	{
        \includegraphics[width=1 in]{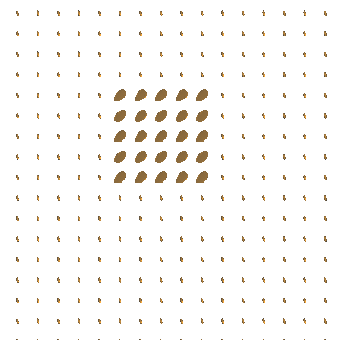}
        \label{fig:third_sub}
    }
	\\
	\subfigure[]
    {
        \includegraphics[width=1 in]{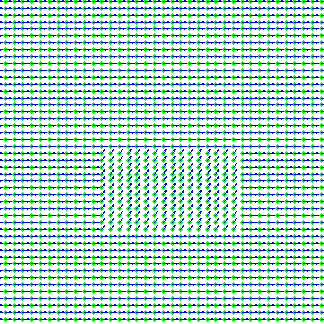}
    }
    \subfigure[]
    {
        \includegraphics[width=1 in]{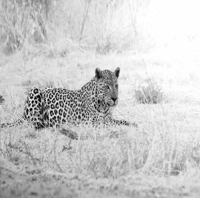}
        \label{fig:second_sub}
    }
    \caption{Example for different type of images. (a) Intensity image. (b) DT-MRI. (c) Vector field with vectors from $S^1$. (d) Texture image with leopard. } \label{fig:Demo}
  \end{figure}
\section{Background}
\label{sec:background}
\subsection{Active contours and Level sets}
In classical active contours \cite{kass}, the user initializes a curve $C(q): [0,1]\rightarrow \Omega \subseteq\mathbb{R}^{2}$ on an intensity image $I:\Omega \rightarrow \Rr$ which evolves and stabilizes on the object boundary. The gradient descent of an energy functional, $E(C)$, given by
\begin{align}
E(C)=\alpha\int_{0}^{1} \| C^{'}(q)\|^{2} dq & + \beta \int_{0}^{1}\| C^{''}(q)\|^{2} dq \\
& - \lambda \int_{0}^{1}\|\nabla I(C(q))\| dq
\end{align}
where $\alpha,\beta$ and $\lambda$ are real positive constants, $C^{'}$ and $C^{''}$ are first and second derivatives of $C$ and $\nabla I$ is the image gradient, gives us a curve evolution equation.  The first two terms are regularizers, while the third term pushes the curve towards the object boundary.\\
Geodesic active contours \cite{kimmel} are an active contour model where the objective function can be interpreted as the length of a curve $C:[0,1] \rightarrow \Rr^2$ in a Riemannian space with metric induced by image intensity. The energy functional for geodesic active contour is given by 
\[
E=\int_{0}^{1} g(\| \nabla I(C(q))\|) \|C'(q)\|\ dq,
\]
where $g:\Rr\rightarrow\Rr$ is a positive monotonically decreasing \emph{edge detector} function. One such choice is $g(s) = \exp\left(-s\right)$. We set $g := g(\|\nabla I\|)$ to make notations simpler. The curve evolution equation that minimizes this energy is given by 
\begin{equation}
\frac{\partial C}{\partial t} = \left(g\kappa - \langle\nabla g, \hat{n}\rangle\right)\hat{n}
\end{equation}
where $\hat{n}$ is the inward unit normal and $\kappa$ is the curvature of the curve $C$.\\
A convenient computational procedure for curve evolution is the level set formulation \cite{osher88,malladi}. Here the curve is embedded in the zero level set of a function $\phi:\mathbb{R}^{2}\rightarrow\mathbb{R}$ which evolves so that the corresponding zero level set evolves according to the desired curve evolution equation. 
For a curve evolution equation of the form $\dfrac{\partial C}{\partial t}=v \hat{n}$, the corresponding level set evolution is $\dfrac{\partial \phi}{\partial t}=v \|\nabla \phi \|$. See Appendix in \cite{kimmel}. In particular the level set evolution for geodesic active contours is given by
\begin{align}
\frac{\partial \phi}{\partial t}=g\cdot \|\nabla\phi\| div\left(\dfrac{\nabla\phi}{\|\nabla\phi\|}\right)+ \nabla g\cdot\nabla\phi.
\end{align}
Another active contour approach was introduced by Chan and Vese \cite{chan} where the energy function was based on regional similarity properties of an object, rather than its edges (image gradient). Suppose that $C$ is the initial curve defined on the domain $\Omega$ of the intensity image $I$. $\Omega$ can be divided into two parts, interior (denoted by $int(C)$) and exterior (denoted by $ext(C)$). 
Let us represent the mean gray value of the region $int(C)$ and $ext(C)$ by $\mu_1$ and $\mu_2$ respectively, then the energy function for which the object boundary is a minimum is given by 
\begin{align}
F_{1}(C)+F_{2}(C) & = \int_{int(C)} (I(x,y)-\mu_1)^{2} dx dy\\
& + \int_{ext(C)}(I(x,y)-\mu_2)^{2} dx dy .
\end{align}
After adding some regularizing terms the energy functional $F(\mu_1,\mu_2,C)$, is given by
\begin{align}
F(\mu_1,\mu_2,C) &=  \mu.Length (C)\; + \;\nu.Area (int(C)) \\
&\;+\; \lambda_{1}\int_{int(C)}(I(x,y)-\mu_1)^{2} dx dy \\
&+ \lambda_{2}\int_{ext(C)}(I(x,y)-\mu_2)^{2} dx dy 
\end{align}
where $\mu\geq 0, \nu \geq 0 , \lambda_{1},\lambda_{2}> 0$ are fixed scalar parameters.
The level set evolution equation is given by
\begin{align}
\frac{\partial \phi}{\partial t}= \delta_{\epsilon}(\phi)\left[\mu \;div\left(\dfrac{\nabla\phi}{\|\nabla\phi\|}\right) - \nu - \lambda_{1}\left(I-\mu_1\right)^{2} +\lambda_{2}\left(I-\mu_2\right)^{2}\right]
\end{align}%
where $\delta_{\epsilon}$ is a smooth approximation of the Dirac delta function, and the level set function $\phi$ is assumed to be negative in the interior region and positive in the exterior region of the curve $C$. A nice survey on active contours and level set implementation can be found in \cite{problems_in_image_processing}.
In this paper, we generalize the two active contour models, the Geodesic active contour and Chan-Vese active contours model to segment \emph{objects} in MVIs. Before describing our active contour model, we will now introduce and define the kind of \emph{images} we focus on in this paper. 
\subsection{Manifold Valued images(MVI)}
In this paper, instead of working with intensity images represented as functions $I:\Omega\subset\mathbb{R}^{2} \rightarrow \mathbb{R}$, we work on images represented as  $I:\Omega \rightarrow M$, where $M$ is a Riemannian manifold\footnote{Manifold is a topological space which is locally Euclidean and a smooth Manifold equipped with a smooth inner product (called Riemannian inner product) on tangent space at every point is called a Riemannian manifold}. In what follows we assume the familiarity with basic concepts from differential geometry like geodesics, Exp map, Log map etc. For the sake of completeness, we define these terms in the Appendix. A thorough introduction and discussion can be found in the textbooks \cite{boothby}, \cite{do}.\\
We present some practical instances of MVI on which we propose to segment "objects" using active contours. Refer to Figure \ref{fig:Demo} to see examples of the following MVIs.
\begin{itemize}
\item[a.] Diffusion tensor magnetic resonance imaging (DT-MRI) produces diffusion tensors corresponding to diffusion of white matter. The diffusion tensors produced are found to be symmetric and positive definite matrices of size $3\times 3$, which forms a Riemannian manifold. A detailed analysis of Diffusion tensor data from DT-MRI can be found in the work by Fletcher and Joshi \cite{fletcher}. In order to display such images in this paper, we represent the positive definite matrix by a planar projection of an ellipsoid with major axis given by its eigenvectors and length of the axis being proportional to the corresponding eigenvalues. 
\item[b.] A variety of vector field images are produced in many application like directional field of wind flow, optical flow \cite{opticalflow}, and others. The manifold in this case could be one of the following: $S^1$, $S^2$, $\mathbb{R}^{2}$, or $\mathbb{R}^{3}$.
\item[c.] Given a texture intensity image $I:\Omega \rightarrow \Rr$, we compute a covariance matrix at every point $(x,y) \in \Omega$ in the following manner. Let $N(x,y)$ denote the $M^2\times 1$ vector containing intensity values of the $M \times M$ neighborhood of the pixel $(x,y)$, arranged in a specific order. Let $C(x,y) = \sum_{(x,y) \in R(x,y)}N(x,y)N(x,y)^T$ be the $M^2 \times M^2$ covariance matrix of $M \times M$ patches in some neighborhood defined by $R(x,y)$, where $N(x,y)^T$ denotes transpose of $N(x,y)$. Covariance matrices are symmetric positive definite\footnote{actually positive semi-definite, but we assume independant random variables giving us positive definiteness.} matrices which form a manifold, denoted by $PD(M^2)$. Thus we get a \emph{manifold valued map} $f_I:\Omega \rightarrow PD(M^2)$.
\end{itemize}
In the last decade or so, a lot of work has been done on manifold data. We mention a few here. Computing nonlinear statistics on such data was proposed by Fletcher and Joshi \cite{fletcher_nonlinear} and by Xavier Pennec \cite{Pennec06}. Weickert and Brox \cite{Weickert02}, Tschumperl\`e and Deriche\cite{Tschumperle03} and Rosman et al.\ \cite{Rosman12} have all proposed regularization schemes for vector valued and/or matrix valued images using PDE's. Tuzel et al.\ \cite{Tuzel08} presented methods using Lie group modeling for tracking objects.\\
In the next section, we describe our active contour model for MVI segmentation.
\section{Adapting active contour model for MVI segmentation}
\label{sec:MVIsegmentation}
We generalize the Geodesic active contours and Chan-Vese active contours for MVIs.
\subsection{Geodesic active contours(GAC)}
In edge dependent active contour models, for example GAC, the curve evolution is made to stop at the object boundary by defining a speed function (or a metric) $g$ that is inversely proportional to an edge detection function which is itself a function of the image gradient (Refer to Section \ref{sec:background}). 
The gradient of a differentiable intensity function $I:\Omega\rightarrow \Rr$ is a vector $v \in \Rr^2$ along which the function $I$ increases the most and whose length $||v||$ is that amount of increase. With Euclidean inner product on $\Rr^2$, the gradient at a point $(p,q) \in \Rr^2$ can be computed as $\nabla I(p,q) = \left(I_x(p,q), I_y(p,q)\right) \in \Rr^2$. Note that in geodesic active contours, the image gradient magnitude (not the image gradient vector itself) plays a significant role: the edge detector function $g$ is a function of the image gradient magnitude. Image gradient magnitude can be interpreted as the maximum rate of change in the value of the function at a point $(p,q) \in \Rr^2$.
\begin{equation}
||\nabla I(p,q)|| = \max_{v\in \Rr^2, ||v|| = 1} \{|DI_{(p,q)}(v)|\} = ||(I_x, I_y)||
\label{eq:gradmag}
\end{equation}
where $DI_{(p,q)}:\Rr^2 \rightarrow \Rr$ is the differential of the function $I$ at point $(p,q)$, and $I_x,I_y$ are evaluated at $(p,q)$. Since this map is linear, the magnitude of maximum increase is the same as the magnitude of maximum decrease in the function value. The function increases the maximum along $\nabla I(p,q)$ and decreases the maximum along $-\nabla I(p,q)$ while the magnitude of rate of \emph{change} in both cases is equal to $||\nabla I(p,q)||$.
We generalize this concept for active contours to work on MVIs. Note that MVI are maps of the kind $I: \Omega\subset \Rr^2 \rightarrow M$. It does not make sense to talk about gradient of such maps, but we define an analogous concept to the gradient magnitude of intensity images for MVIs. The differential of the MVI function\footnote{In what follows, we assume that $I$ is differentiable in the sense described in the book\cite{boothby}. The definition of Differential of a smooth map between manifolds is given in the Appendix \ref{appendix}.} $I$ at point $(p,q) \in \Rr^2$ is given by
\begin{align}
& DI_{(p,q)} : T_{(p,q)}\Rr^2(\simeq \Rr^2) \rightarrow T_{I(p,q)}M \\
& DI_{(p,q)}(a,b) = aI_x + bI_y \in T_{I(p,q)}M,
\label{eq:MVIdiff}
\end{align}
where $T_{I(p,q)}M$ is the tangent space to $M$ at the point $I(p,q)$, $I_x = DI_{(p,q)}(1,0)$ and $I_y = DI_{(p,q)}(0,1)$. Moreover
\begin{equation}
||DI_{(p,q)}(a,b)||_{I(p,q)} = ||aI_x + bI_y||_{I(p,q)}
\end{equation}
where $||.||_{I(p,q)}$ is the Riemannian norm on the tangent space $T_{I(p,q)}M$. We define the gradient magnitude of $I$ at $(p,q)$ (denoted by $\delta_M I(p,q)$) as the maximum possible value of the norm of the differential: 
\begin{equation}
\delta_M I(p,q) = \max_{||(a,b)||_{\Rr^2} = 1}||aI_x + bI_y||_{I(p,q)}
\end{equation}
Let the inner product on $T_{I(p,q)}M$ be defined as 
\begin{equation}
\langle u,v \rangle_{I(p,q)} = u^tG_{(p,q)} v
\end{equation}
where $G_{(p,q)}$ is a symmetric positive definite matrix that varies smoothly over $M$. Let
\begin{equation}
A_{pq} = 
\left[ 
\begin{array}{cc}
I_x^tG_{(p,q)}I_x    &  I_x^tG_{(p,q)}I_y  \\
I_y^tG_{(p,q)}I_x    &  I_y^tG_{(p,q)}I_y
\end{array} \right].
\end{equation}
Then it can be shown that 
\begin{equation}
\delta_M I(p,q) = \sqrt{\lambda_{\max}\left(A_{pq}\right)},
\label{eq:MVIgradmag}
\end{equation}
where $\lambda_{\max}\left(A_{pq}\right)$ is the maximum singular value of the matrix $A_{pq}$. Finally we explain how we compute $I_x(p,q)$ and $I_y(p,q)$ on discretized MVIs. 
In case of intensity images, the following approximation is frequently used:
\begin{align}
I_x(p,q)\ \simeq\ & I(p+1,q)-I(p,q)\\
I_y(p,q)\ \simeq\ & I(p,q+1)-I(p,q).
\end{align}
Subtraction in vector space can be re-interpreted on Riemannian manifolds as the Riemannian Log map\cite{pennec08}. Using this, the analogous approximations to $I_x,I_y$ for MVIs $I:\Omega \rightarrow M$, are given by
\begin{align}
I_x(p,q) \simeq I(p+1,q)-I(p,q) = Log_{I(p,q)}\left(I(p+1,q)\right)\\
I_y(p,q) \simeq I(p,q+1)-I(p,q) = Log_{I(p,q)}\left(I(p,q+1)\right),
\end{align}
where $Log_a(b) \in T_aM$ is the Riemannian Log map at point $a \in M$ for point $b \in M$.
Having defined the required gradient magnitude for MVIs, the level set evolution equation corresponding to geodesic active contours is given as
\begin{equation}
\frac{\partial \phi}{\partial t} =  \|\nabla\phi\|\ div\left(g \cdot\;\frac{\nabla\phi}{\|\nabla\phi\|}\right),
\label{eq:MVIGAC}
\end{equation}
where
\begin{equation}
g(\delta_M I(p,q)) =  \frac{1}{1+\delta_{M}I(p,q)}.
\end{equation}
\subsection{Chan-Vese active contour model}
In Chan-Vese active contour model the energy function is 
\begin{align}
F(\mu_1,\mu_2,C) &= \mu. \; Length (C)\; + \;\nu. \; Area (int(C))\\
&\;+\; \lambda_{1}\int_{int(C)} (u_{0}(x,y)-\mu_1)^{2} dx dy \\ &+ \lambda_{2}\int_{ext(C)} (u_{0}(x,y)-\mu_2)^{2} dx dy 
\end{align}
where $\mu_1$ and $\mu_2$ are mean over the interior and exterior of the contour $C$, respectively.\\
For MVIs, $\mu_{i}, i = 1,2$ are the intrinsic means of the manifold data in the interior and exterior of the curve $C$. Intrinsic mean of a collection of points $x_{1},...,x_{n}\in M$ is the minimizer of the sum of squared Riemannian distances from each of the given points:
\begin{equation}
\mu=argmin_{x\in M}\sum_{i=1}^{n} d(x,x_{i})^2
\end{equation}
where $d(.,.)$ is the Riemannian distance on $M$ and is a generalization of the Euclidean distance $d(x,y) = ||x-y||$ to Riemannian manifolds \cite{pennec08}. This is computed using a gradient descent approach. For a detailed explanation refer to \cite{fletcher_nonlinear}, and refer to Section \ref{sec:expts} for a computational algorithm to compute the intrinsic mean. 
The corresponding energy function for Chan-Vese active contours on MVI is thus given by
\begin{align}
\nonumber F(\mu_{1},\mu_{2},C) &= \mu. \; Length (C)\; + \;\nu. \; Area (int(C)) \\
\nonumber & +\; \lambda_{1}\int_{int(C)} d(I(x,y),\mu_{1})^{2} dx dy \\
& + \lambda_{2}\int_{ext(C)} d(I(x,y),\mu_{2})^{2} dx dy 
\end{align}
where $\mu_{1}$ and $\mu_{2}$ are intrinsic means of manifold points lying in the regions $int(C)$ and $ext(C)$  respectively. The level set evolution equation comes out to be
\begin{align}
\nonumber \frac{\partial \phi}{\partial t} = \delta_{\epsilon}(\phi) \left[ \mu \;div\left( \frac{\nabla\phi}{\|\nabla\phi\|}\right) -\nu -\right. & \lambda_{1}  d(I(x,y),\mu_{1})^{2} \nonumber\\
& \left. +\lambda_{2}  d(I(x,y),\mu_{2})^{2}\right],
\end{align}
where, following the Chan-Vese model, we assume that the level set function $\phi$ is negative in the interior and positive in the exterior of the curve $C$.
In the next section, we implement both GAC and Chan-Vese active contour model on several examples of MVIs. We also provide computational algorithms for computing Riemannian Exp map, Log map and intrinsic mean on the corresponding manifolds. The Riemannian Exp map is the inverse of Riemannian Log map, and intuitively can be interpreted as vector addition in Riemannian manifolds \cite{pennec08}.
\section{Experiments}
\label{sec:expts}
Our Active contour model requires computing the Exp map, Log map and intrinsic mean for each manifold. For short definitions of these terms refer to the Appendix \ref{appendix}. A summary of algorithms to compute these maps on $S^2$ ($2$-D sphere), $SO(3)$($3$-D rotation matrices) and $PD(3)$ manifolds are provided next, while detailed derivations can be found in one or more of these papers: \cite{fletcher_symmetricspace,fletcher_nonlinear,exp_so3,Moakher11}.
\subsection{Computation on different manifolds}
\begin{enumerate}
\item $S^2$ 
\begin{enumerate}
\item Exp map
\begin{equation}
Exp_{p}(v) = \cos(\| v\|)p+ \sin(\| v\|)\dfrac{v}{\| v\|}
\end{equation}
where $v\in T_{p}(S^2)$ and $p \in S^2$.
\item Log map 
\begin{equation}
Log_{p}(q) = \dfrac{\theta}{sin \theta}(q-p cos(\theta)),\quad \theta= \cos^{-1}\langle p,q\rangle,
\end{equation}
where $p,q \in S^2$.
\end{enumerate}
\item $SO(3)$ 
\begin{enumerate}
\item Exp map
\begin{align}
Exp_{p}(v) = & p\ \exp(v)\\
           = &  p\left(I+ \dfrac{sin(\| \widehat{v}\|)}{\| \widehat{v}\|}v + \dfrac{1-cos(\| \widehat{v}\|)}{\| \widehat{v}\|^{2}}v^{2}\right)
\end{align}
where $v\in T_p(SO(3))=so(3) $ and $\widehat{v}$ is a representation of $v$ in $\mathbb{R}^{3}$.
\item Log map
\begin{align}
Log_{p}(q) = &\ Log(p^Tq)\\
           = &\ \frac{\theta}{2 \sin \theta}
           \begin{bmatrix}
-R(1,2)+R(2,1) \\
R(3,1)-R(1,3) \\
-R(2,3)+R(3,2)
\end{bmatrix}
\end{align}
where $p,q \in SO(3)$, $R=p^Tq$ and $\theta = cos^{-1}\left(\frac{Trace(R)-1}{2}\right)$.
\end{enumerate}
\item $PD(n)$
\begin{enumerate}
\item Exp map \\
\textbf{Input :} Initial point $p\in PD(n)$ \\
 Tangent vector $X\in Sym(n)$\\
 \textbf{Output :} {$Exp_{p}(X)$}\\
   Let $p=u\Lambda u^{T} (u\in SO(n), \Lambda$ diagonal)\\
   $g=u\sqrt{\Lambda}$\\
   $Y=g^{-1}X(g^{-1})^{T}$\\
   Let $Y=v\Sigma v^{T} (v\in SO(n), \Sigma$ diagonal)\\
   $Exp_{p}(X)=(gv)exp(\Sigma)(gv)^{T}$\\
\item Log map \\
 \textbf{Input :} Initial point $p\in PD(n)$\\
  End point $q\in PD(n)$\\
 \textbf{Output :} $Log_{p}(q)$ \\
  Let $p=u\Lambda u^{T} (u\in SO(n), \Lambda$ diagonal)\\
  $g=u\sqrt{\Lambda}$\\
  $y=g^{-1}q(g^{-1})^{T}$\\
  Let $y=v\Sigma v^{T} (v\in SO(n), \Sigma$ diagonal)\\
  $Log_{p}(q)=(gv)log(\Sigma)(gv)^{T}$\\
\end{enumerate}
\end{enumerate}
We have used the following algorithm to find the intrinsic mean on all manifolds, details of which can be found in \cite{fletcher_nonlinear}.
\RestyleAlgo{boxruled}
 \begin{algorithm}
 \label{mean}
 \KwIn{$x_{1},x_{2},...,x_{n}\in M$}
 \KwOut{$\mu \in M$, the intrinsic mean}
  \quad $\mu_{0}=x_{1}$\\
  \quad Do \\
  \quad $\Delta \mu = \frac{1}{N}\Sigma_{i=1}^{N} Log_{\mu_{j}}x_{i}$\\
  \quad $\mu_{j+1}=Exp_{\mu_{j}}(\Delta \mu)$\\
  \quad While $\Vert\Delta \mu\Vert > \epsilon$\\
  \caption{Intrinsic Mean on a manifold} 
\end{algorithm}
With all the required computational machinery set, we next demonstrate our segmentation results for both Geodesic active contours and Chan-Vese active contours on various manifold valued images.
\subsection{Results}
\begin{figure}
    \centering
    \subfigure[]
    {
        \includegraphics[width=1 in]{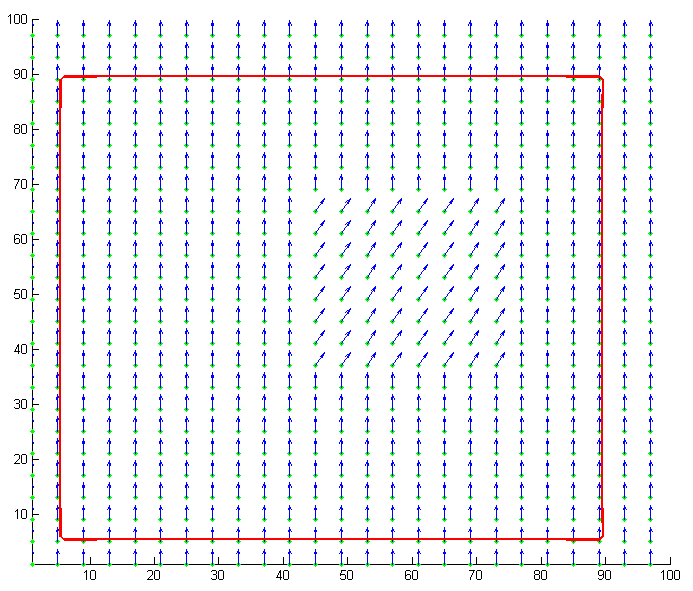}
        \label{s1_1}
    }
    \subfigure[]
    {
        \includegraphics[width=1 in]{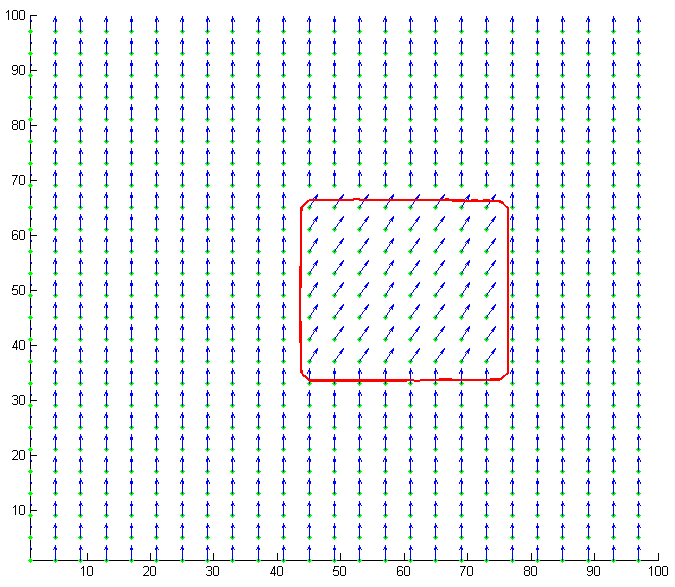}
        \label{s1_gac}
    }
    \subfigure[]
    {
        \includegraphics[width=1 in]{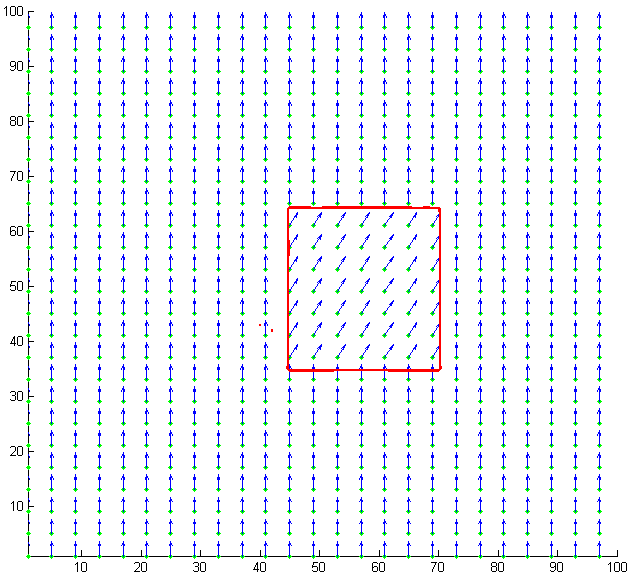}
        \label{s1_chan}
    }
    \\
     \subfigure[]
    {
        \includegraphics[width=1 in]{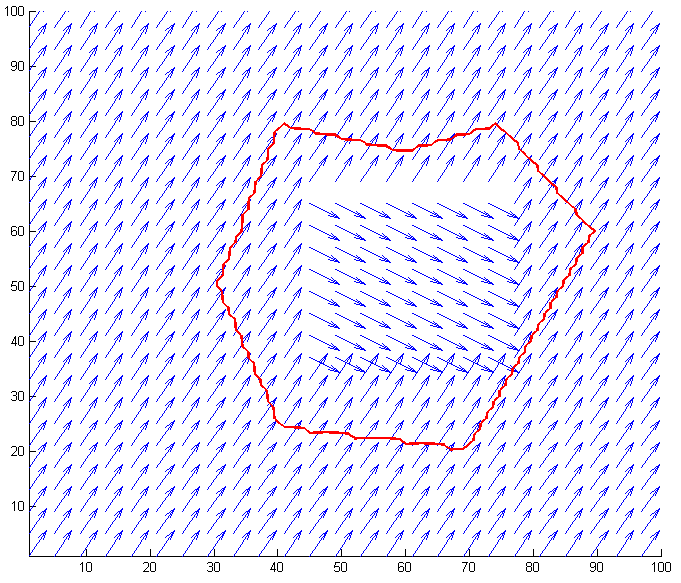}
        \label{s2_1}
    }
    \subfigure[]
    {
        \includegraphics[width=1 in]{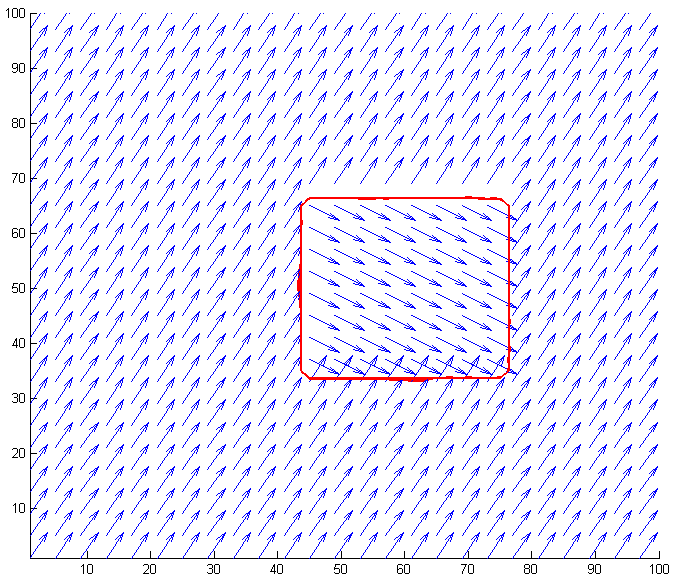}
        \label{s2_gac}
    }
    \subfigure[]
    {
        \includegraphics[width=1 in]{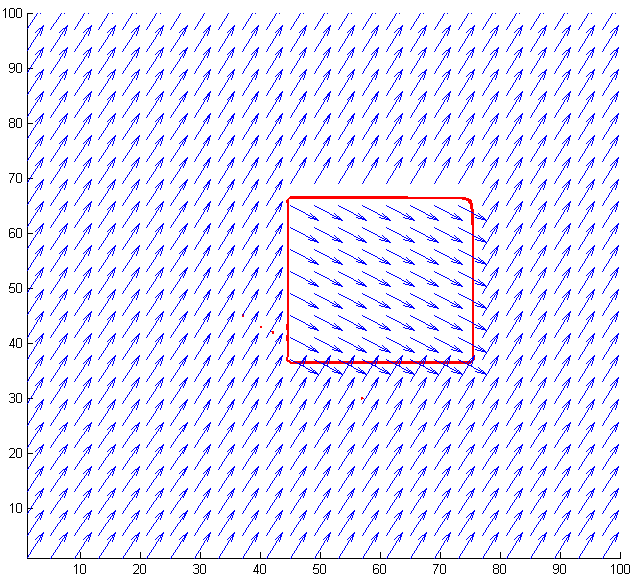}
        \label{s2_chan}
    }
    \\
      \subfigure[]
    {
        \includegraphics[width=1 in]{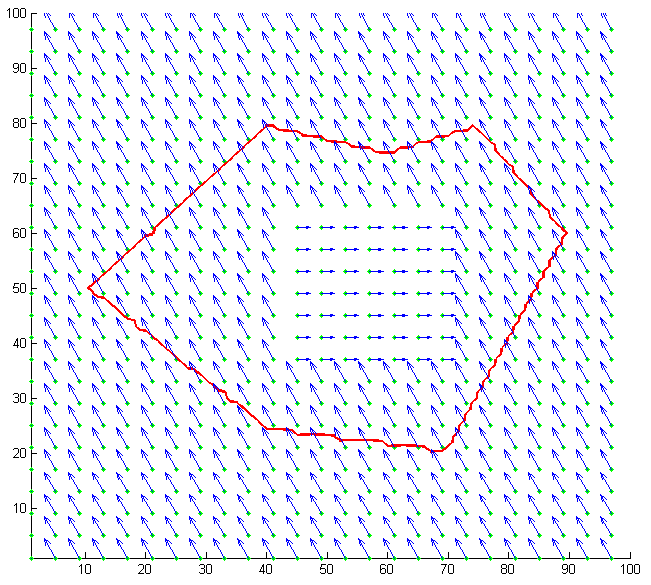}
        \label{so3_1}
    }
    \subfigure[]
    {
        \includegraphics[width=1 in]{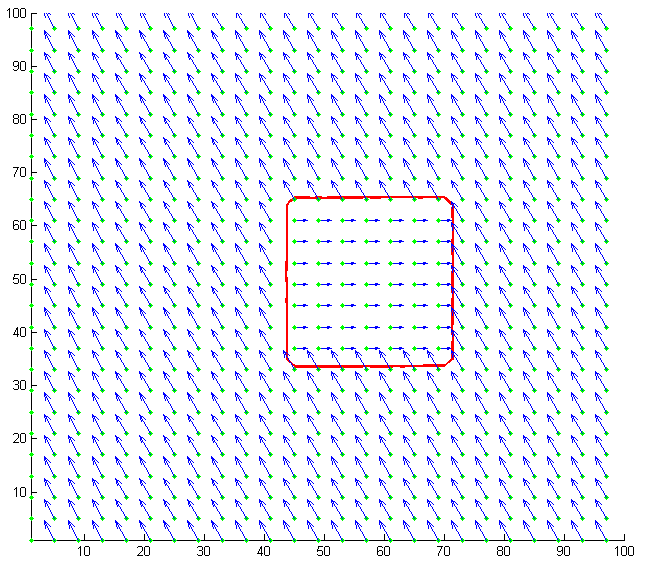}
        \label{so3_gac}
    }
    \subfigure[]
    {
        \includegraphics[width=1 in]{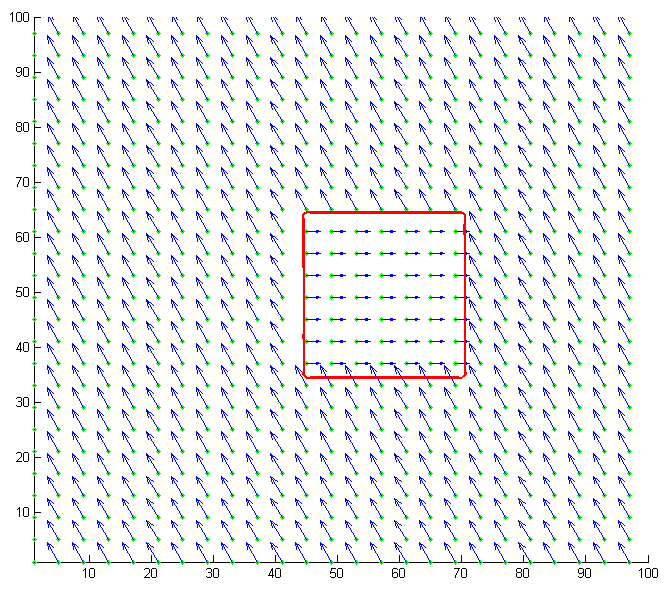}
        \label{so3_chan}
    }
    \\
     \subfigure[]
    {
        \includegraphics[width=1 in]{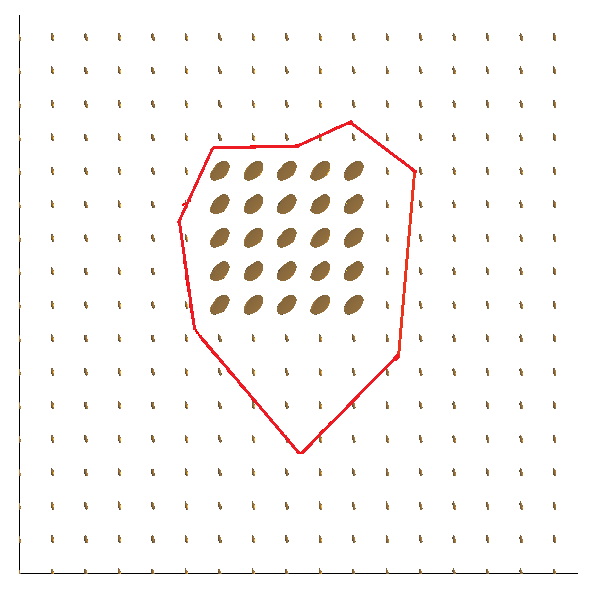}
        \label{pd3_1}
    }
    \subfigure[]
    {
        \includegraphics[width=1 in]{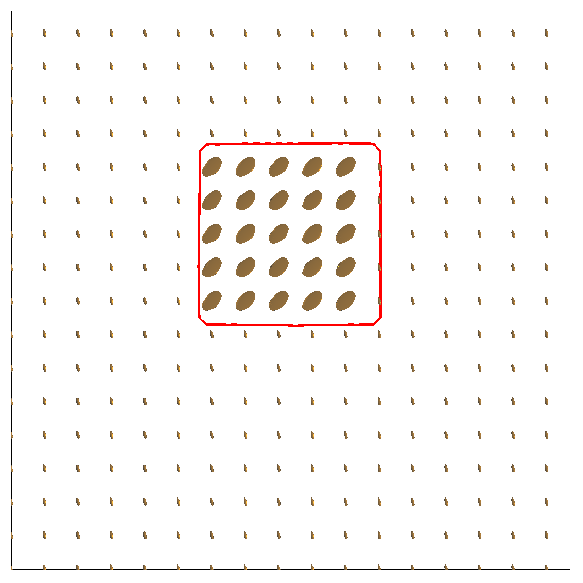}
        \label{pd3_gac}
    }
    \subfigure[]
    {
        \includegraphics[width=1 in]{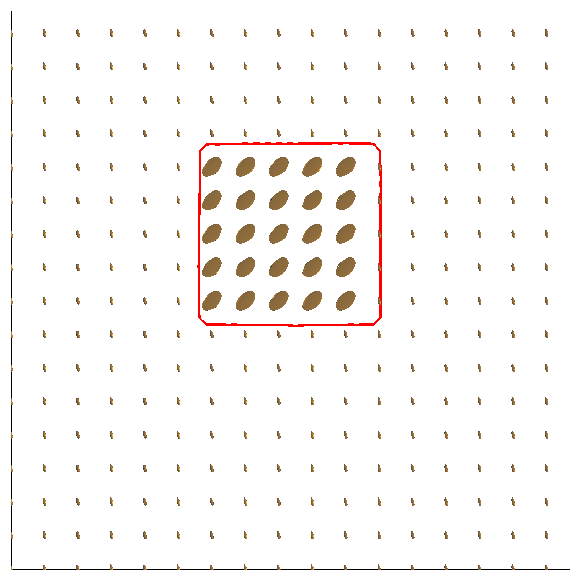}
        \label{pd3_chan}
    }
         \caption{Segmentation results on different types of manifold. (left column) Original images with manifolds $S^1$, $S^2$, $SO(3)$ and $PD(3)$ in top to bottom order, with initial contour. (center column) Corresponding geodesic active contour segmentation results. (right column) Corresponding Chan-Vese active contour segmentation results.} \label{Results on demo MVIs}
\end{figure}

\begin{figure}
    \centering
    \subfigure[]
    {
        \includegraphics[width=1 in]{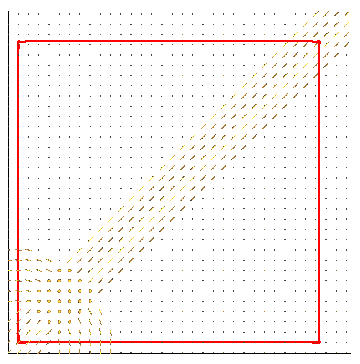}
        \label{fig:DT-MRI Real}
    }
    \subfigure[]
    {
        \includegraphics[width=1 in]{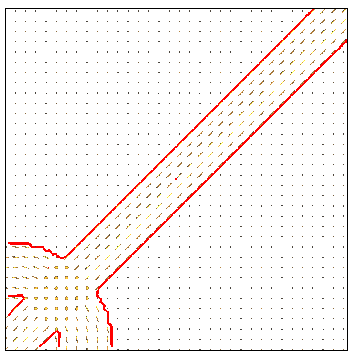}
        \label{fig:DT-MRI Real 2}
    }
    \caption{Results on Real DT-MRI data.}
    \label{Result on DT_MRI Real data}
\end{figure}

\begin{figure}
    \centering
    \subfigure[]
    {
        \includegraphics[width=1 in]{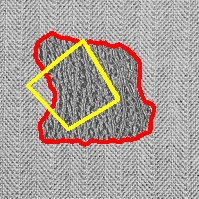}
        \label{texure1}
    }
    \\
    \subfigure[]
    {
        \includegraphics[width=1 in]{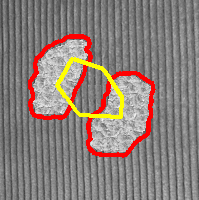}
        \label{texture2}
    }
    \subfigure[]
    {
        \includegraphics[width=1 in]{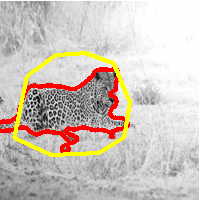}
        \label{tetxure3}
    }
    \caption{Segmentation results on different texture images. (a) and (b) show the results on small scale texture images, (c) shows the result on a leopard image.} \label{Result on texture images}
\end{figure}

Results on $S^1$, $S^2$, $SO(3)$ and $PD(3)$ manifold valued images for Geodesic active contours as well as Chan-Vese active contours are shown in Figure \ref{Results on demo MVIs}. Result on Real DT-MRI data\footnote{from \url{http://www.cise.ufl.edu/~abarmpou/lab/fanDTasia/}} is shown in Figure \ref{Result on DT_MRI Real data}.
Texture segmentation problem can be posed as an MVI segmentation problem, as already explained in Section \ref{sec:background}. We obtain a $PD(M^2)$-valued image from a texture image over which our algorithm successfully segments different textures\footnote{from \url{http://sipi.usc.edu} and \url{http://www.nada.kth.se/cvap/databases/kth-tips}} as shown in Figure \ref{Result on texture images}. Since texture boundaries are not defined based on gray-value image edges, the geodesic active contour model does not yield appropriate segmentation results. The results shown are for Chan-Vese active contour model. We have used a $5 \times 5$ covariance matrix to characterize a texture and it is computed over a neighborhood of $13 \times 13$ pixels. 
\section{Conclusion and future scope}
In this paper we have generalized the active contours for MVI segmentation. We provide several such examples which can be dealt under our general framework. The drawbacks and benefits of the proposed model are same as those of the usual active contour models. As a computer vision application we pose the texture segmentation problem as an MVI segmentation problem and demonstrate some texture segmentation results using our algorithm. We take a neighborhood of size $5\times 5$ pixels in an intensity image to form covariance matrices over a larger neighborhood of size $13 \times 13$ at every pixel to get a $PD(25)$-valued image . This fixes the scale of features in which we are interested. Some textures may need multi scale information to be properly characterized. Simply increasing the neighborhood size is not going to help in dealing with large scale texture as it may prevent proper localization of the texture boundary. One needs to incorporate a mechanism that automatically detects the scale of the given texture. \\
The computational cost for the model extended from Chan-Vese active contour is high since one needs to compute the Riemannian Exp and Log map for every manifold value in the image. This cost will of course vary from manifold to manifold. The texture segmentation algorithm is computationally very expensive: To generate results on images of size $150 \times 150$ pixels, it took about $30$ minutes on a $2$ GHz Intel laptop. This is primarily due to repeated Exp and Log map computation on $PD(25)$ manifold. One may use multi-grid approaches to speed up the evolution.
\bibliographystyle{plain}
\bibliography{JMIV2013_paper}
\section{Appendix}
\label{appendix}
We give some basic definitions from Differential geometry required for our paper. For a thorough explanation, we refer the reader to the books \cite{boothby}, \cite{do}.
\begin{enumerate}
\item Differentiable Manifolds:\\
  A differentiable manifold of dimension $n$ is a set $M$ and a family
  of injective mappings ${\mathord{\mathcal T}}$ = $\{x_i: U_i \subset {\mathord{\mathbb R}}^n \to M\}$ of open
  sets $U_i$ of ${\mathord{\mathbb R}}^n$ into $M$ such that
  \begin{itemize}
  \item $\bigcup _i x_i(U_i) = M$, i.e. the open sets cover $M$.
  \item for any pair $i,j$ with $x_i(U_i)\bigcap
    x_j(U_j) = W \neq \phi$, the mapping $x_j^{-1} \circ
    x_i$ is differentiable.
  \item The family $\mathord{\mathcal T}$ is maximal, which means that if $(y,V)$, $y:V\subset
    {\mathord{\mathbb R}}^n\to M$ is such that: for each element of $\mathord{\mathcal T}$, $(x_i,U_i)$ with
    $x_i(U_i)\cap y(V)\not = 0$ implies that $y^{-1}\circ x_i$ is a diffeomorphism,
    then in fact $(y,V)\in\mathord{\mathcal T}$.
  \end{itemize}
\item Differential of a smooth map between differential manifolds:\\
Let $F:M\rightarrow N$ be a smooth map between two differentiable manifolds $M$ and $N$. Given a point $p \in M$, the differential of $F$ at $p$ is a linear map 
\[
dF_p: T_pM \rightarrow T_{F(p)}M
\]  
from the tangent space of $M$ at $p$ to the tangent space of $N$ at $F(p)$. 
\item Riemannian Metric:\\
  A Riemannian metric on a manifold $M$ is a correspondence
  which associates to each point $p\in M$ an inner product
  $\langle -,-\rangle_p$ on the tangent space $T_pM$, which
  varies smoothly. In terms of local coordinates, the metric at
  each point $x$ is given by a matrix, $g_{ij} = \langle X_i,X_j\rangle_x$,
  where $X_i,X_j$ are tangent vectors to $M$ at $x$,
  and it varies smoothly with $x$. A \textit{Geodesic curve} is
  a local minimizer of arc-length computed with a Riemannian
  metric.
\item Geodesics:\\
 A parameterized curve $\gamma:I\rightarrow M$ is a geodesic if
 $\frac{D}{dt}\left(\frac{d\gamma}{dt}\right) = 0, \forall t \in I$, where $\frac{D}{dt}$ is called the \emph{Covariant derivative} and intuitively is the orthogonal projection of the usual derivative to the tangent space. Geodesics are also local minimizers of arclength.
\item Exponential map:\\
  The exponential map is a map $Exp : TM \to M$, that maps $v \in
  T_qM$ for $q \in M$, to a point on $M$ obtained by going out the
  length equal to $|v|$, starting from $q$, along a geodesic which
  passes through $q$ with velocity equal to $\frac{v}{|v|}$. The
  geodesic starting at $q$ with initial velocity $t$ can thus be
  parametrized as
  $$
  t\mapsto Exp_q(tv).
  $$
\item Log map: For $\tilde{q}$ in a sufficiently small neighborhood of
  $q$, the length minimizing curve joining $q$ and $\tilde{q}$ is
  unique as well. Given $q$ and $\tilde{q}$, the direction in which to
  travel geodesically from $q$ in order to reach $\tilde{q}$ is given
  by the result of the logarithm map $Log_q(\tilde{q})$. We get the
  corresponding geodesic as the curve
  $t\mapsto Exp_q(tLog_q\tilde{q})$. In other words, $Log$ is the
  inverse of $Exp$ in the neighborhood.
\end{enumerate}
\end{document}